\documentclass[10pt,twocolumn,letterpaper]{article}

\usepackage[pagenumbers]{cvpr}      

\usepackage[pdf]{pstricks}
\usepackage{graphicx}
\usepackage{amsmath}
\usepackage{amsfonts}
\usepackage{amssymb}
\usepackage{booktabs}
\usepackage{xcolor}
\usepackage{multirow}
\usepackage{bbm}
\def\1{\mathbbm{1}}
\definecolor{yellowcrop}{RGB}{255, 255, 0}

\usepackage[pagebackref,breaklinks,colorlinks]{hyperref}

\usepackage[capitalize]{cleveref}
\crefname{section}{Sec.}{Secs.}
\Crefname{section}{Section}{Sections}
\Crefname{table}{Table}{Tables}
\crefname{table}{Tab.}{Tabs.}

\def\confName{CVPR}
\def\confYear{2022}

\begin{document}
\title{Towards Universal Texture Synthesis by Combining Texton Broadcasting with Noise Injection in StyleGAN-2}

\author{Jue Lin\\
Northwestern University, ECE Department\\
2145 Sheridan Rd, Evanston, IL 60208, USA\\
{\tt\small jue.lin@u.northwestern.edu}
\and
Gaurav Sharma\\
University of Rochester, ECE Department\\
Rochester, NY\\
{\tt\small gaurav.sharma@rochester.edu}
\and
Thrasyvoulos N. Pappas\\
Northwestern University, ECE Department\\
2145 Sheridan Rd, Evanston, IL 60208, USA\\
{\tt\small t-pappas@northwestern.edu}
}

\twocolumn[{%
\renewcommand\twocolumn[1][]{#1}%
\maketitle
\begin{center}
    \centering
    \captionsetup{type=figure}
\begingroup
\setlength{\tabcolsep}{1pt}
\renewcommand{\arraystretch}{0.5}
\begin{tabular}{llllll}
\includegraphics[width=0.108\textwidth]{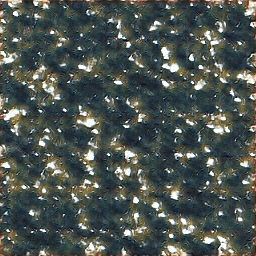}&\multirow{2}{*}[0.673in]{\includegraphics[width=0.216\textwidth]{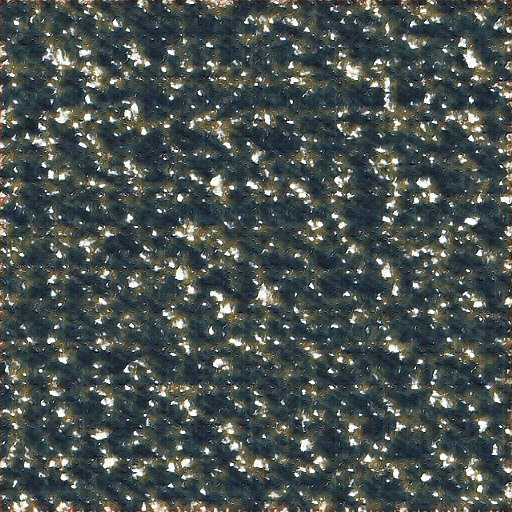}}&\includegraphics[width=0.108\textwidth]{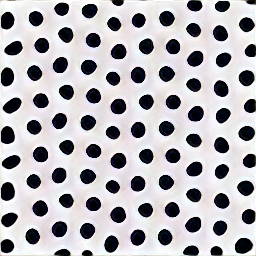}&\multirow{2}{*}[0.673in]{\includegraphics[width=0.216\textwidth]{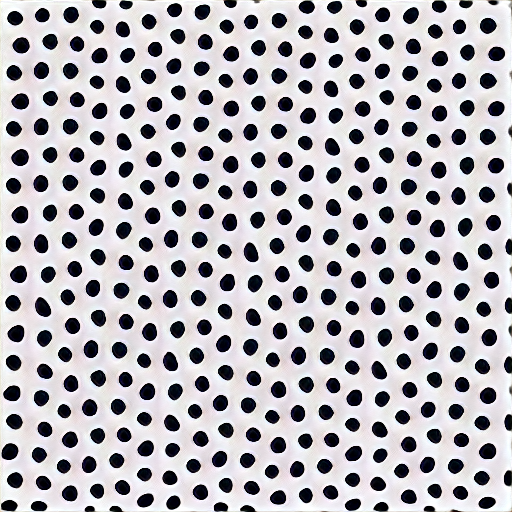}}&\includegraphics[width=0.108\textwidth]{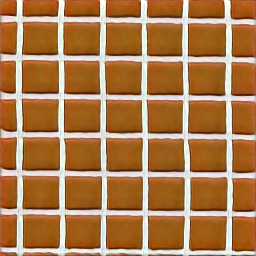}&\multirow{2}{*}[0.673in]{\includegraphics[width=0.216\textwidth]{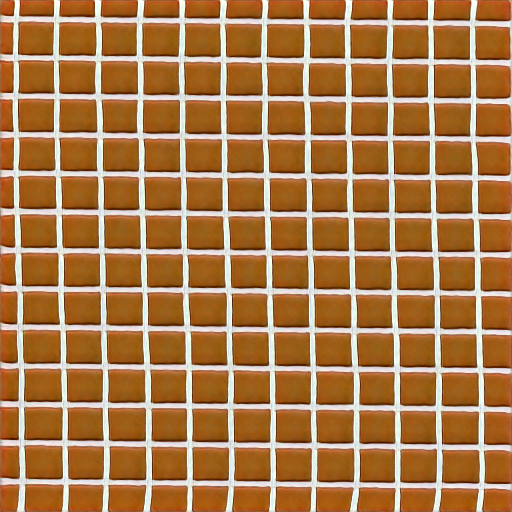}}\\
\includegraphics[width=0.108\textwidth]{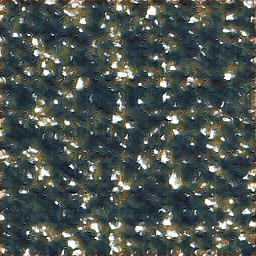}&&\includegraphics[width=0.108\textwidth]{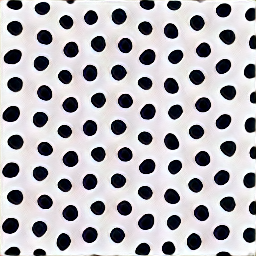}&&\includegraphics[width=0.108\textwidth]{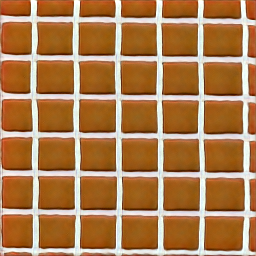}&
\end{tabular}
\endgroup
\captionof{figure}{Textures synthesized at $256\times256$ and $512\times512$ pixel resolutions. The proposed model is trained on $256\times256$ texture crops, and can produce textures ranging from stochastic to structured with variable resolutions, conditioned on the latent representation.}
\end{center}%
}]
\maketitle

\begin{abstract}
\vspace*{-5pt}
We present a new approach for universal texture synthesis by incorporating a multi-scale texton broadcasting module in the StyleGAN-2 framework. The texton broadcasting module introduces an inductive bias, enabling generation of broader range of textures, from those with regular structures to completely stochastic ones. To train and evaluate the proposed approach, we construct a comprehensive high-resolution dataset that captures the diversity of natural textures as well as stochastic variations within each perceptually uniform texture. Experimental results demonstrate that the proposed approach yields significantly better quality textures than the state of the art. The ultimate goal of this work is a comprehensive understanding of texture space.
\vspace*{-20pt}
\end{abstract}
\section{Introduction}
\label{sec:intro}
Texture is an important visual attribute for human perception and computer vision, as it provides critical information for material appearance, understanding, and characterization \cite{adelson_hvei01}.
Texture understanding, and texture analysis/synthesis in particular, is important for a variety of applications, including image analysis and compression, computer graphics, virtual reality, and human-computer interaction.
The study of texture analysis/synthesis must take into account the stochastic nature of texture and human perception, which is the ultimate judge of texture quality.
Accordingly, a number of authors have proposed algorithms for texture analysis/synthesis that are based on multiscale frequency decompositions, which have been used to model early visual processing in the brain \cite{cano88,porat89,heeger95b,portilla96,zhu96,debonet97,portilla00}.
On the other hand, the stochastic nature of texture necessitates a statistical approach for texture analysis. 
The most complete parametric approach for texture analysis/synthesis has been proposed by Portilla and Simoncelli \cite{portilla00}, who developed a statistical model for synthesizing a broad set of textures based on a steerable filter decomposition.
Even though their goal was to provide a universal statistical model that parametrizes the space of visual textures, it falls short of successfully modeling all textures.
Thus, the complete mathematical and perceptual characterization of texture remains an open problem.

The resurgence of neural networks has stimulated broad interest in both academia and industry, promising to push the frontiers in a wide variety of research areas. One of the most successful models is the generative adversarial network (GAN), which has yielded impressive results in numerous applications, such as generation of human faces, anime characters, objects and scenes, image-to-image translation, image super-resolution, and inpainting.
However, the problem of texture modeling has not received as much attention.  The focus of this work is on utilizing the GAN framework for texture synthesis, and ultimately, a more complete mathematical and perceptual characterization of the space of visual textures. We present a new approach for universal texture synthesis that introduces a multiscale texton broadcasting module in the StyleGAN-2 framework, which enables the generation of a wide variety of textures, both regular and stochastic.

A generally accepted definition of visual texture is an image that is spatially homogeneous and usually contains repeated elements, often with random variations in position, orientation, and color \cite{portilla_simoncelli}. The repeated elements in a texture are commonly referred to as {\it textons,} a term introduced by Bela Julesz, one of the pioneers of texture analysis and perception \cite{julesz81}. Texture appearance can range from completely regular periodic structure to completely random variations, and typically consists of both periodic structure and stochastic variations. However, in our experiments we found that the failure of the StyleGAN-2 framework to capture the periodic aspect of textures manifests itself in two ways: first, the trained model generates disproportionately fewer periodic textures when randomly sampled from the latent space; and second, once we identify a latent vector corresponding to a periodic texture, injecting the model with different samples of multi-scale noise cannot produce distinguishable texture crops, that is, the synthesized periodic structure is often ``anchored'' in a fixed location.  We will analyze both of these shortcomings and will show that they can be mitigated by the introduction of the texton broadcasting module. 

The main contributions of this work are the following:
\vspace*{-4pt}
\begin{itemize}
\setlength\itemsep{-2pt}
\item We propose a novel multi-scale texton broadcasting module for the SyleGAN-2 that, in combination with the noise injection, provides appropriate inductive bias to enable universal high-quality texture synthesis, spanning from regular to completely stochastic ones.
\item For effective training and evaluation, we create a comprehensive dataset of high-resolution textures, representative of the diversity of natural textures as well as the variations within individual textures.
\item We introduce an intuitive measure to quantify how well a network models distribution within individual texture, and we emphasize the importance of the often overlooked multi-scale noise injection and our texton broadcasting in this regard. 
\item We demonstrate that unlike conventional analysis-synthesis techniques, traversing in our trained latent space exhibits smooth transitions between homogeneous textures instead of incoherent spatial mixtures.
\end{itemize}

\section{Related work}
\subsection{Texture analysis and synthesis}
\label{subsec:texture_analysis_and_synthesis}

As we discussed, traditional parametric approaches for texture analysis/synthesis have been primarily based on subband decompositions. Heeger and Bergen \cite{heeger95b} match the histograms of a steerable filter decomposition to achieve impressive texture synthesis results; however, their approach is limited to stochastic textures. Portilla and Simoncelli \cite{portilla00} developed a more elaborate model that relies on a wide variety of subband statistics for synthesizing a much broader set of textures.  Their goal is to parametrize the space of visual textures based on a universal statistical model. Parametric approaches based on Markov Random Fields (MRFs) have also shown great potential for texture synthesis \cite{Levina:TextureSynMRF:AS06, Paget:TextureSynMRF:TIP98}. Nonparametric approaches for texture synthesis build larger textures from seed patches \cite{efros_iccv99, kwatra_siggraph03}; however, this does not involve any texture modeling.

Deep learning-based approaches for texture analysis/synthesis are fairly diversified. 
Gatys \etal \cite{Gatys:TextureSynthCNN:NIPS2015} use a pretrained VGG-19 network to extract features from a given texture and then, starting with a random image, synthesize another texture that matches the Gram-matrix representation of the original texture.  Ulyanov \etal \cite{Ulyanov:TexNet:ICML2016, Ulyanov:imp_texturenet:CVPR2017} use a fast feed-forward generative network to achieve similar performance.
Li \etal \cite{Li:DivTex:CVPR2017} further develop a feed-forward generative network to synthesize multiple diverse textures. 
Implicit neural representation (INR) methods \cite{Chen:ImplicitDecode:CVPR2019, Mescheder:OccuNet:CVPR2019, Park:DeepSDF:CVPR2019, Sitzmann:INR:NIPS2020} can be used to predict RGB values of a texture using raw pixel coordinates as input \cite{Henzler:3DTexture:CVPR2020, Oechsle:TexField:ICCV2019, Portenier:GramGAN:NIPS2020}. 
However, most non-GAN methods perform texture reconstruction without a meaningful latent space representation. 
For that we turn to GANs.

\subsection{Generative adversarial networks}
Goodfellow \etal \cite{Goodfellow:GAN:NIPS2014} introduced an adversarial formulation for training a generative model, whereby a second discriminator network provides feedback by determining whether a generated image comes from the actual data distribution or not. 
The WGAN \cite{Arjovsky:WGAN:ICML2017} uses the Wasserstein (or earth mover) distance between probability distributions, which improves stability of learning and alleviates mode collapse. 
Further improvements come from alternative formulations of Lipschitz continuity for the WGAN, e.g., gradient penalty \cite{Gulrajani:WGAN-GP:NIPS2017} and spectral normalization \cite{Miyato:WGAN-SN:ICLR2018}. 

The application of GANs to texture synthesis was introduced by Jetchev \etal \cite{Jetchev:SpatialGAN:NIPS2016W}, who  proposed the spatial GAN for synthesis of textures of arbitrary size.
However, like Gatys \etal \cite{Gatys:TextureSynthCNN:NIPS2015}, the functionality of the spatial GAN is limited to producing equivalent textures, that is, it generates one model per texture.
The periodic spatial GAN (PSGAN) by Bergmann \etal \cite{Bergmann:PSGan:ICML2017} represents the first attempt to learn a latent space that is capable of generating periodic textures
by injecting a periodic pattern (with a random phase term) at the lowest resolution of the generator network.  However, they trained on a very small dataset and, as we will show below, the quality of the resulting textures is mixed. 
 
\subsection{StyleGAN models}
Building on the progressive-GAN \cite{Zhang:PGan:NIPS2019}, StyleGAN, proposed by Karras \etal \cite{Karras:StyleGAN:CVPR2020}, introduces an intermediate latent space, which is used to adjust the style of the image at each convolution layer, and also adds explicit noise injection at each layer.  This allows the disentanglement of global features (like pose, face shape and human identity) and local stochastic variations (like hair and skin texture). 
StyleGAN-2 \cite{Karras:StyleGAN2:CVPR2020} was proposed to address some noticeable blob-like visual artifacts in StyleGAN, by redesigning the normalization and eliminating progressive training.

Following the Karras \etal work, which was applied to faces, objects, or scenes, one line of research sought to interpret the latent space induced by StyleGAN-like models. Built upon loss functions containing location information or pretrained attribute classifiers, an input image can be inverted into a latent code $\mathbf{z}$, and visual property manipulation can be achieved via navigating in the latent space \cite{Abdal:Image2StyleGAN:ICCV2019, Shen:FaceEdit:CVPR2020, Shen:InterFace:PAMI2020}.  However, due to lack of an equivalent model, a proper GAN inversion technique is still missing for textures. Although loss functions for faces, objects, or scenes have been proposed, they rely on pixel-to-pixel correspondences, which are ill-suited for textures because of their stochastic nature.
 
Another line of research sought to investigate the impact of the internal components of the StyleGAN models. Xu \etal \cite{Xu:PositionEncode:CVPR2021} found that zero-padding implicitly encodes location, which works for faces, objects, or scenes but is not desirable for textures. Choi \etal\cite{Choi:MSPE:ICCV2021} addressed the spatial bias in StyleGAN-2 by adding sinusoidal embeddings, commonly used in transformers \cite{Vaswani:Transformer:NIPS2017, Gehring:CSSL:ICML2017, Kolesnikov:VIT:ICLR2021}. 

Finally, the recently proposed StyleGAN-3 \cite{Karras:StyleGAN3:NIPS2021} replaces the bottom tensors with Fourier features \cite{Tancik:FourFeat:NIPS2020} and designs operators to enhance translation/rotation equivariance. It is interesting to note that, for better performance on human faces, StyleGAN-3 removes the noise injection, which we found to be highly beneficial for textures. 

\section{Method}
\subsection{Preliminaries}
We denote by $\Omega=\{\Omega_1, \Omega_2,\cdots\}$ the set of all textures. Intuitively, each texture $\Omega_i$ has a set of basic elements or {\it textons}\cite{julesz60} (\eg, a brick in a wall). We can aggregate textons from $\Omega$ and establish a {\it universal texton codebook} $T_{\Omega}$\cite{leung_malik_ijcv01}. Any texture $\Omega_i$ can be represented as a spatial repetition of textons drawn from $T_{\Omega}$, and the selected textons are adjusted to fit certain properties (\eg, shapes). Moreover, stochastic variation is often present within each $\Omega_i$ (\eg, layout of bricks). We can therefore model texture distribution in two parts: {\it inter}-texture distribution $P_{\Omega_i\sim\Omega}(\Omega_i)$ for distinct textures, and {\it intra}-texture distribution $P_{\mathbf{I}\sim\Omega_i}(\mathbf{I}_i)$ conditioned on the same texture $\Omega_i$, where we assume $\Omega_i$ is sufficiently large and $\mathbf{I}$ is a random crop from $\Omega_i$. 

Conceptually, a StyleGAN-2 generator $G_{\boldsymbol\theta}$ can approximate both $P_{\Omega_i\sim\Omega}(\Omega_i)$ and $P_{\mathbf{I}\sim\Omega_i}(\mathbf{I})$ with $P_{\mathbf{z}, \mathbf{n}}(G_{\boldsymbol\theta}(\mathbf{z}, \mathbf{n}))$ and $P_{\mathbf{n} | \mathbf{z}}(G_{\boldsymbol\theta}(\mathbf{z}, \mathbf{n})|\,\mathbf{z})$ respectively, where $\mathbf{z}$ is drawn from a D-dimensional normal distribution and $\mathbf{n}$ is the multi-scale spatial noise. Due to randomness in the recurrence of textons, most textures exhibit {\it stochasticity} as well as {\it periodicity}. The stochasticity is well captured by $\mathbf{n}$, producing subtle changes in textured regions, \eg, hair and skin. However, our experiments show that StyleGAN-2 generator is biased towards synthesizing stochastic textures, even though the regular ones occupy a comparable portion in the training set, which we refer to as {\it inter-texture mode collapse}. Moreover, the injected noise $\mathbf{n}$ is inadequate for rendering spatial shifts of the periodic structures, which we refer to as {\it intra-texture mode collapse}. We empirically found that the first problem is relatively easier to alleviate. 

\begin{figure}[t]
     \centering
     \begin{subfigure}[b]{0.23\textwidth}
         \centering
         \includegraphics[width=1\textwidth]{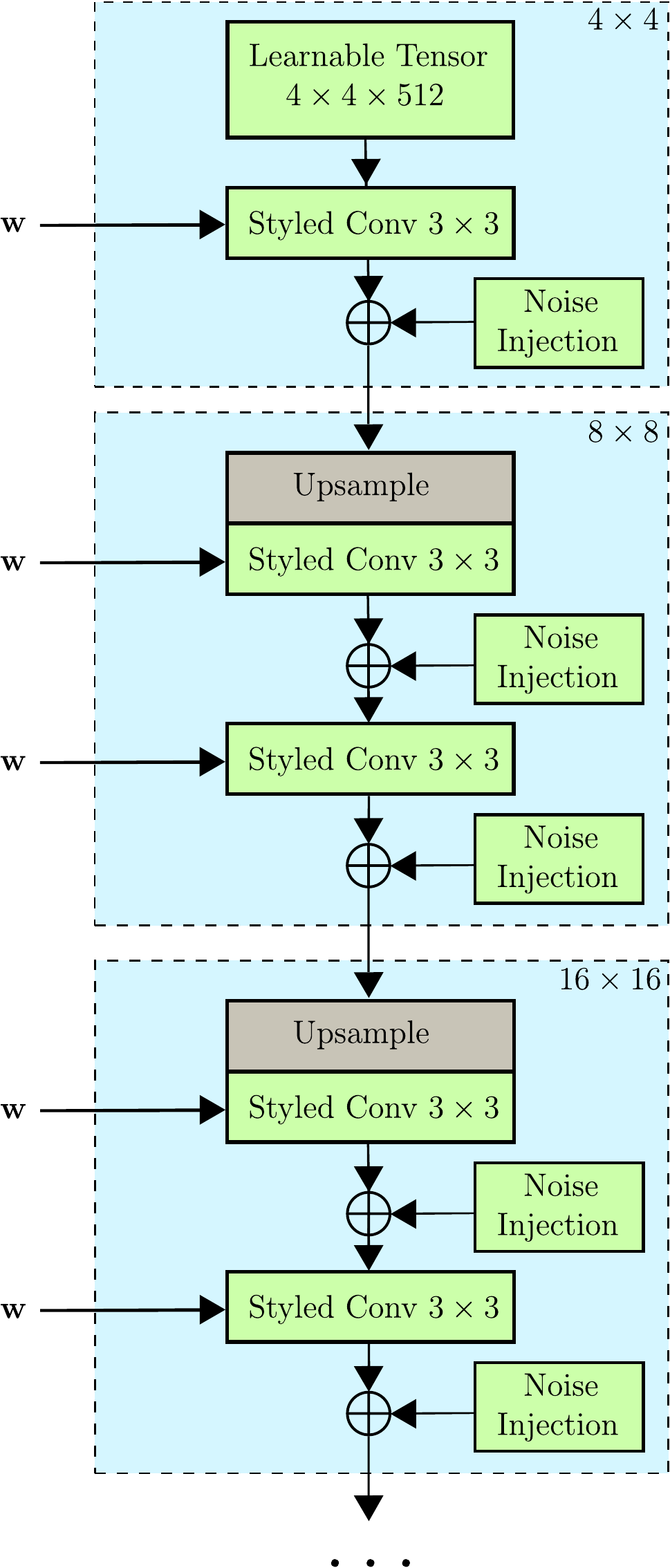}
         \caption{StyleGAN-2 generator}
         \label{fig:stylegan2}
     \end{subfigure}
     \hfill
     \begin{subfigure}[b]{0.23\textwidth}
         \centering
         \includegraphics[width=1\textwidth]{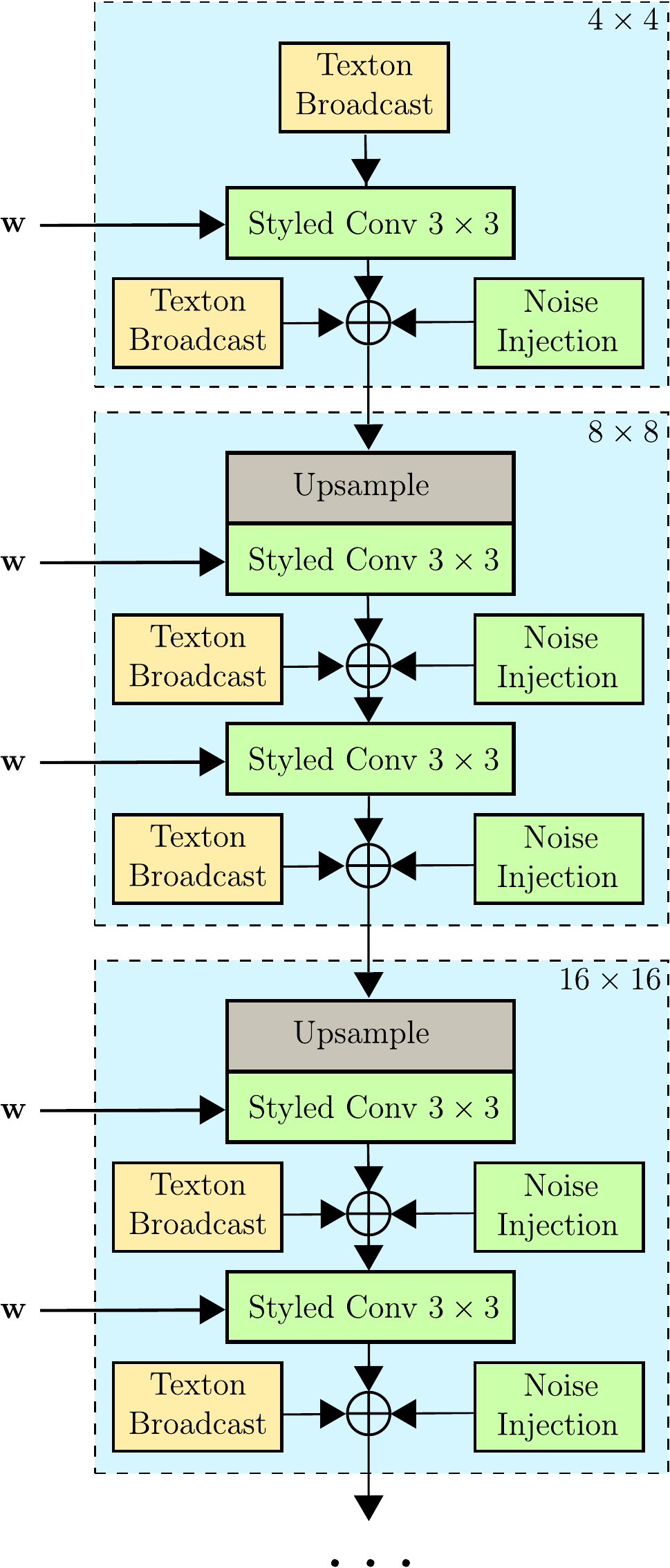}
         \caption{Multi-Scale Texton Broadcast}
         \label{fig:msTextureGen}
     \end{subfigure}
        \caption{Comparison between StyleGAN-2 and our proposed modifications. All feature maps within the same blue area, enclosed with dashed line, share the same spatial resolution, indicated on the upper right corner. The "$\oplus$" is an element-wise sum.}
        \label{fig:modelCompare}
\vspace*{-6pt}
\end{figure}

Nonetheless, the intra-texture mode collapse suggests a strong entanglement between spatial location and latent space. Such entanglement is arguably acceptable for images in other domains, \eg, face positioned in the center, but the spatial layout of textons should be stochastic. Most operations in StyleGAN-2 (see Fig.~\ref{fig:modelCompare}) do not explicitly encode spatial information, \eg, convolutions, upsampling. We have identified the bottom $512\times4\times4$ tensor and zero-padding in coarse layers as the causes of spatial anchoring of visible structures (see Fig.~\ref{fig:exp_intra_mc_stylegan2}).

\subsection{Texton broadcasting module}
\begin{figure}[t]
\centering
\includegraphics[width=0.5\textwidth]{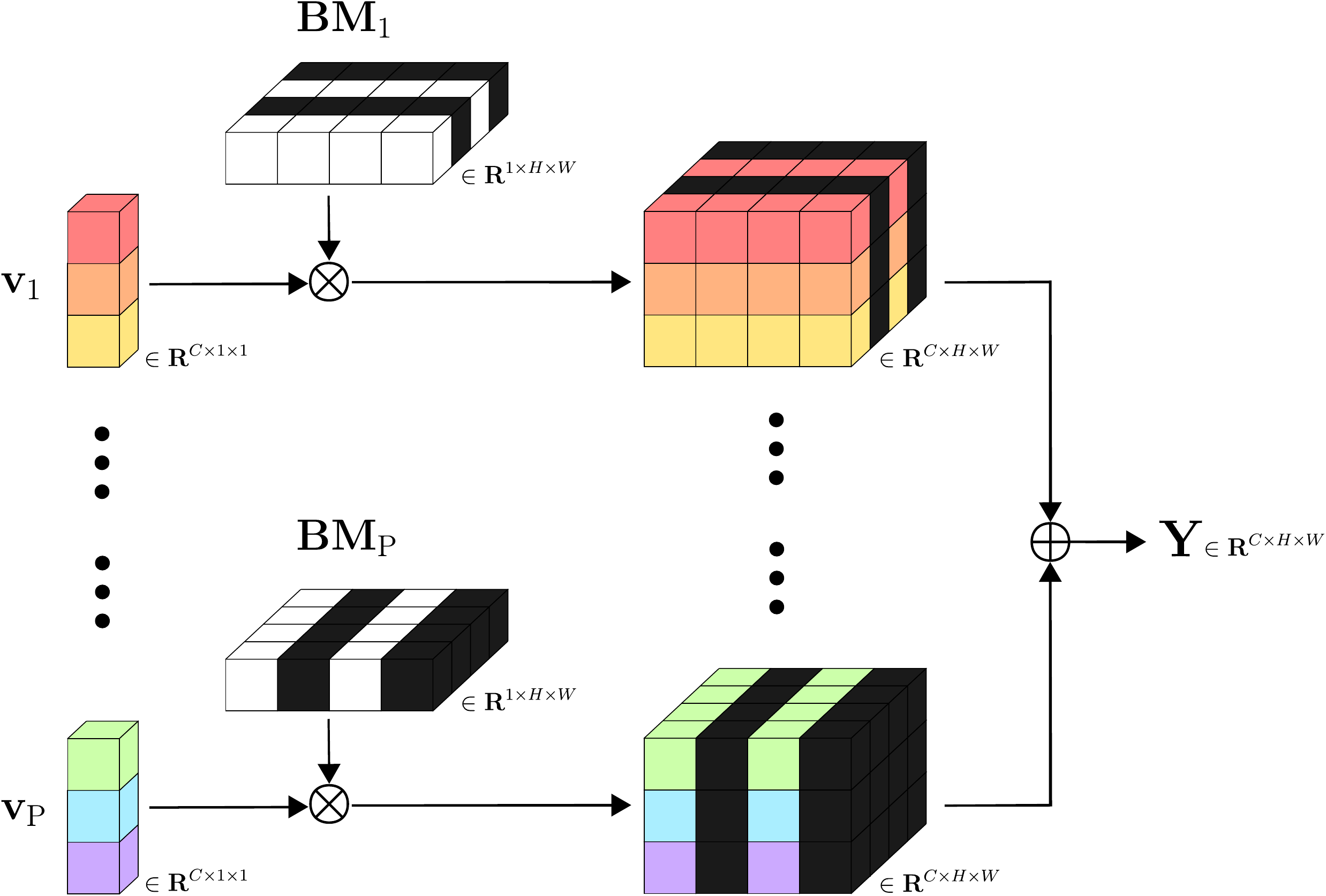}
\caption{Mechanism of Texton Broadcasting. Each cubelet represents a scalar, the value of which is indicated by its color. For ease of illustraion, we assume a black cubelet $=0$, a white cubelet $=1$, and other colors are arbitrary. The "$\otimes$" duplicates $\textbf{v}_i$ spatially and $\textbf{BM}_i$ along channel dimension, then an element-wise multiplication follows. The "$\oplus$" is an element-wise sum.}
\label{fig:textonBroadcast}
\end{figure}

To capture the periodic nature of textures, we design a {\it texton broadcasting (TB)} module that simulates the spatial repetition of physical textons, as illustrated in Fig.~\ref{fig:textonBroadcast}. First, a trainable texton $\mathbf{v}_i$ is replicated along spatial dimensions. Then the intensity of each $\mathbf{v}_i$ is modulated with respect to a {\it broadcast map (BM)}, modelled as a 2D sinusoidal wave:
\begin{equation}\label{eq:2d_wave}
    \begin{aligned}
       \text{BM}_i(h, w) = \textbf{A}_i \sin(2\pi \varsigma(\mathbf{f}_i)^T\begin{bmatrix}h\\w\end{bmatrix}+\boldsymbol\varphi_i+\Delta)+\textbf{B}_i
    \end{aligned}  
\end{equation}
where $\forall\,i\in\{1,2,\ldots,P\},\ \textbf{f}_i=[f_{ih}, f_{iw}]^T$, $\boldsymbol\varphi_i$, $\textbf{A}_i$ and $\textbf{B}_i$ represent the frequency, initial phase, amplitude, and offset of the 2D sine, all of which are trainable parameters, and $P$ denotes the total number of textons in a TB module. We use $[h, w]^T\in\{1,2, \ldots, H\}\times\{1,2, \ldots, W\}$ as the spatial coordinate vector; $H$ and $W$ can be dynamically sized but are fixed during training. We use an element-wise sigmoid function $\varsigma(\cdot)$ to map $[f_{ih}, f_{iw}]^T$ into the interval $(0, 1)$, since discrete-time frequency is periodic in $\omega$ with period $2\pi$, \ie, $\sin(\omega n)=\sin((\omega+2\pi)n), \forall n\in\mathbb{Z}$. Note that $\Delta$ is uniformly sampled from $[0, 2\pi)$ to simulate random shift, and is shared among all $\text{BM}_i$ within the same module to have a unified phase control. The output $\textbf{Y}$ of a module is the sum across all broadcast/modulated $\mathbf{v}_i$:
\begin{equation}\label{eq:output}
    \begin{aligned}
       \textbf{Y}(c, h, w)=\sum_{i=1}^{P} \textbf{v}_i(c)\otimes\text{BM}_i(h, w)
    \end{aligned}  
\end{equation}
where $\otimes$ combines replication of all textons $\mathbf{v}_i$ with the modulation by $\text{BM}_i$, as illustrated in Fig.~\ref{fig:textonBroadcast}. Note that the use of random phase $\Delta$ is critical, otherwise the module is just another deterministic spatial anchor.

\subsection{Multi-scale texton broadcasting}
By replacing the bottom $512\times4\times4$ tensor with a TB module, the new model is now capable of producing images of variable sizes. However, the zero padding at the bottom layers still causes a residual spatial anchoring effect when synthesizing images of higher resolution. To mitigate this issue, we couple each noise injection (NI) module with a TB module (see Fig.~\ref{fig:msTextureGen}), upto and including layers with spatial size of $64\times64$. Such hierarchical placement of NI and TB is aligned with the multi-scale nature of textures. The spatial size $H$ and $W$ of each TB module is configured to match the resolution at its corresponding layer except the bottom one, which is set to $4\times4$ to produce a final $256\times256$ image compatible with the discriminator during training\footnote{Code can be found \href{https://github.com/JueLin/textureSynthesis-stylegan2-pytorch}{here}}.

\subsection{Training objective functions}
The inter-texture mode collapse is in fact closely related to the general definition of mode collapse in the literature, where a model yields only a few distinguishable images. We adopt the Wasserstein distance as the loss function, and impose a gradient penalty \cite{Gulrajani:WGAN-GP:NIPS2017} to enforce the Lipschitz continuity on the discriminator network (or critic) $D_{\boldsymbol\phi}$, parametrized by $\boldsymbol\phi$. The losses for $D_{\boldsymbol\phi}$ and $G_{\boldsymbol\theta}$ are given as:
\begin{align}
\mathcal{L}(\boldsymbol\phi)=&\ \mathbb{E}_{(\textbf{z},\textbf{n},\Delta_G)}\{D_{\boldsymbol\phi}(G_{\boldsymbol\theta}(\mathbf{z}, \mathbf{n}, \Delta_G))\}-\notag\\
&\ \ \ \ E_{\Omega_i\sim\Omega}\{\mathbb{E}_{\textbf{I}|\Omega_i}\{D_{\boldsymbol\phi}(\textbf{I})\}\}\label{eq:loss_D}\\
\mathcal{L}(\boldsymbol\theta)=&-\mathbb{E}_{(\textbf{z},\textbf{n},\Delta_G)}\{D_{\boldsymbol\phi}(G_{\boldsymbol\theta}(\mathbf{z}, \mathbf{n}, \Delta_G))\}\label{eq:loss_G}
\end{align}
where $\mathbf{z}$ and $\mathbf{n}$ are drawn from normal distribution, $\Delta_G$ denotes the set of all $\Delta$ injected across $G_{\boldsymbol\theta}$. Note that in (\ref{eq:loss_D}), multiple crops $\textbf{I}$ are sampled from the same texture $\Omega_i$, which helps $D_{\boldsymbol\phi}$ learn intra-texture distribution explicitly.

\begin{figure*}[t]
\centering
\begin{subfigure}[b]{1\textwidth}
\includegraphics[width=0.12\textwidth]{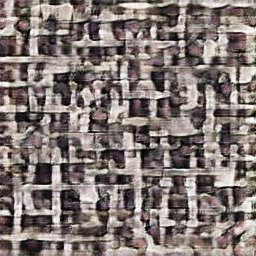} 
\includegraphics[width=0.12\textwidth]{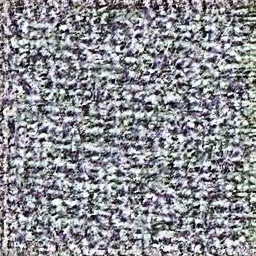} 
\includegraphics[width=0.12\textwidth]{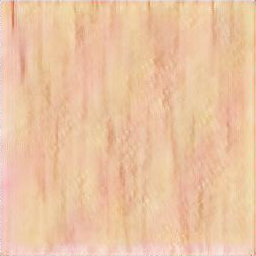} 
\includegraphics[width=0.12\textwidth]{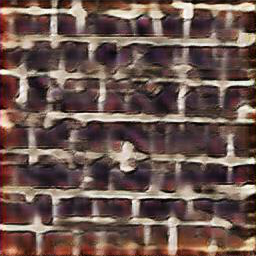} 
\includegraphics[width=0.12\textwidth]{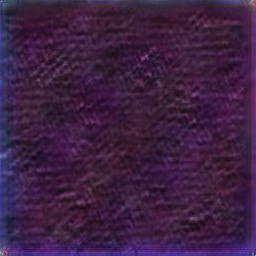} 
\includegraphics[width=0.12\textwidth]{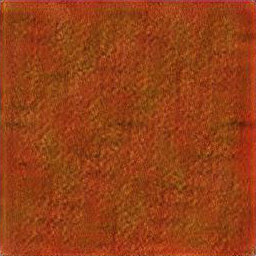} 
\includegraphics[width=0.12\textwidth]{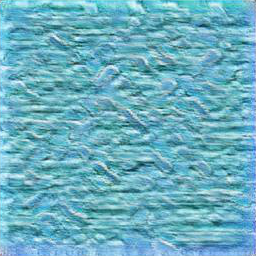} 
\includegraphics[width=0.12\textwidth]{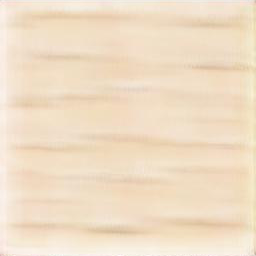}
 
\includegraphics[width=0.12\textwidth]{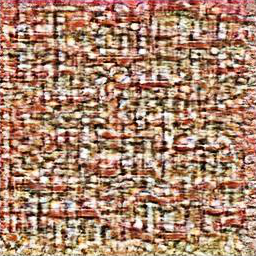} 
\includegraphics[width=0.12\textwidth]{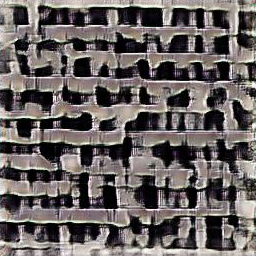} 
\includegraphics[width=0.12\textwidth]{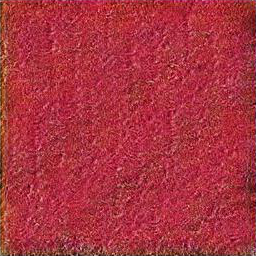} 
\includegraphics[width=0.12\textwidth]{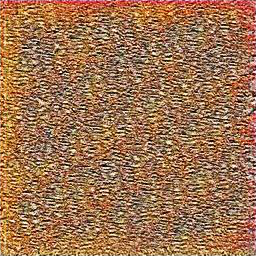} 
\includegraphics[width=0.12\textwidth]{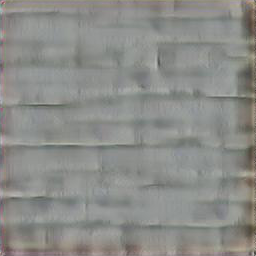} 
\includegraphics[width=0.12\textwidth]{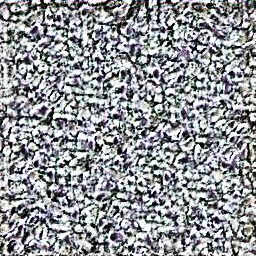} 
\includegraphics[width=0.12\textwidth]{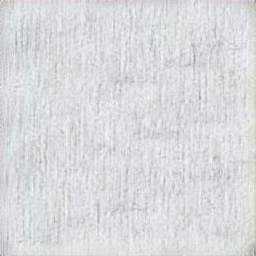} 
\includegraphics[width=0.12\textwidth]{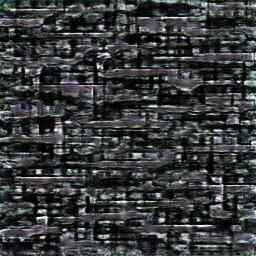}
 \caption{PSGAN: suboptimal texture reconstructions}
\end{subfigure}

\begin{subfigure}[b]{1\textwidth}
\includegraphics[width=0.12\textwidth]{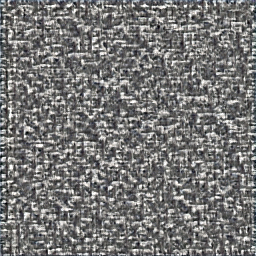} 
\includegraphics[width=0.12\textwidth]{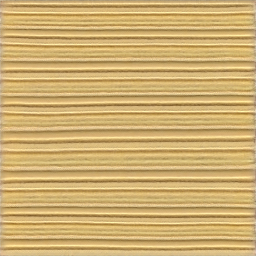} 
\includegraphics[width=0.12\textwidth]{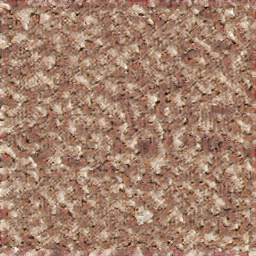} 
\includegraphics[width=0.12\textwidth]{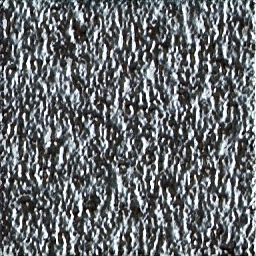} 
\includegraphics[width=0.12\textwidth]{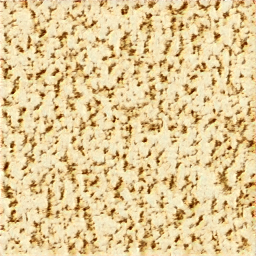} 
\includegraphics[width=0.12\textwidth]{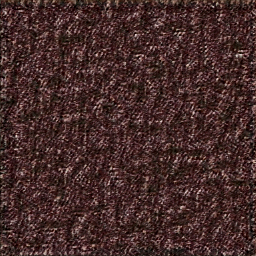} 
\includegraphics[width=0.12\textwidth]{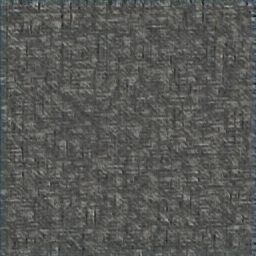} 
\includegraphics[width=0.12\textwidth]{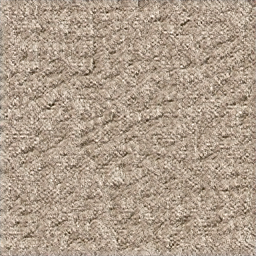}

\includegraphics[width=0.12\textwidth]{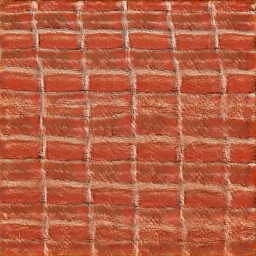} 
\includegraphics[width=0.12\textwidth]{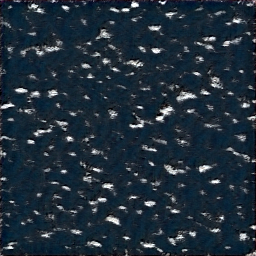} 
\includegraphics[width=0.12\textwidth]{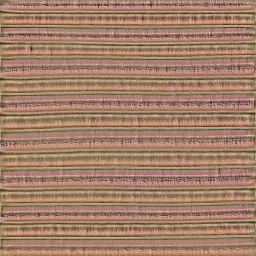} 
\includegraphics[width=0.12\textwidth]{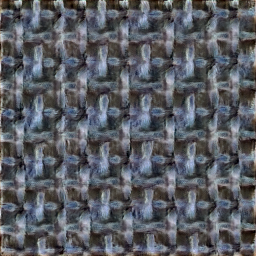} 
\includegraphics[width=0.12\textwidth]{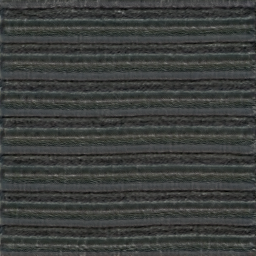} 
\includegraphics[width=0.12\textwidth]{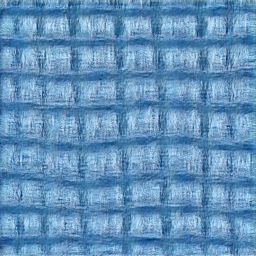} 
\includegraphics[width=0.12\textwidth]{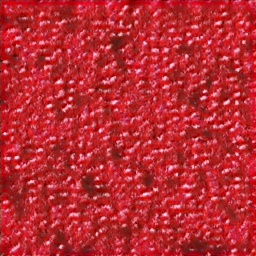} 
\includegraphics[width=0.12\textwidth]{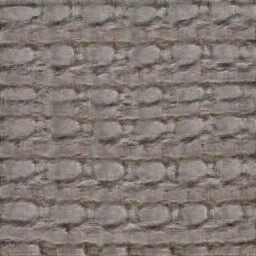}
\caption{StyleGAN-2: visually competitive, but severe intra-texture mode collapse, discussed in later sections}
\end{subfigure}

\begin{subfigure}[b]{1\textwidth}
\includegraphics[width=0.12\textwidth]{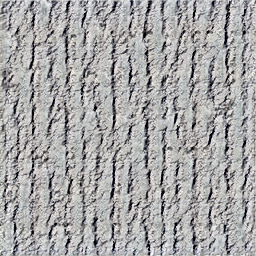}
\includegraphics[width=0.12\textwidth]{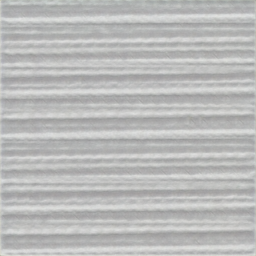}
\includegraphics[width=0.12\textwidth]{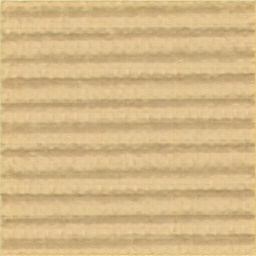}
\includegraphics[width=0.12\textwidth]{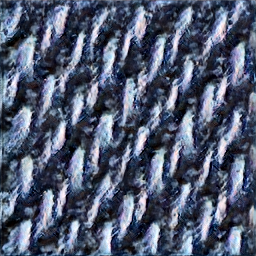}
\includegraphics[width=0.12\textwidth]{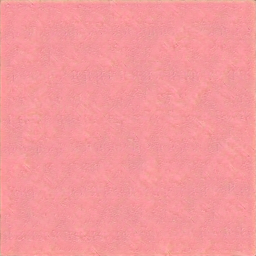}
\includegraphics[width=0.12\textwidth]{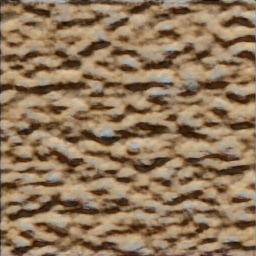}
\includegraphics[width=0.12\textwidth]{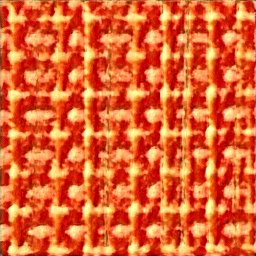}
\includegraphics[width=0.12\textwidth]{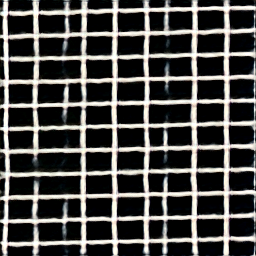}

\includegraphics[width=0.12\textwidth]{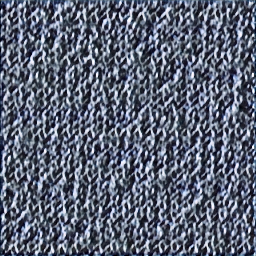}
\includegraphics[width=0.12\textwidth]{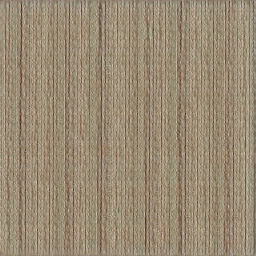}
\includegraphics[width=0.12\textwidth]{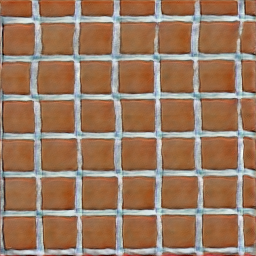}
\includegraphics[width=0.12\textwidth]{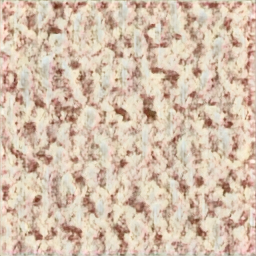}
\includegraphics[width=0.12\textwidth]{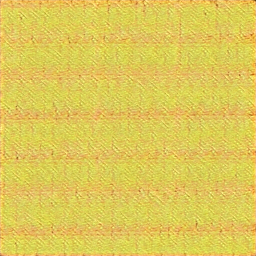}
\includegraphics[width=0.12\textwidth]{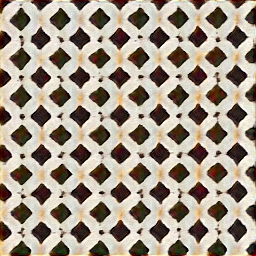}
\includegraphics[width=0.12\textwidth]{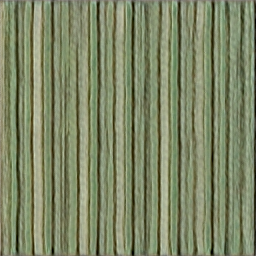}
\includegraphics[width=0.12\textwidth]{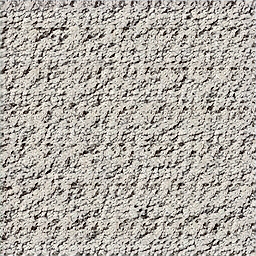}
 \caption{Proposed approach: well defined textures, good balance between stochastic and structured/periodic textures}
\end{subfigure}
\vspace*{-14pt}
\caption{Lists of textures generated by models trained on our dataset}
\label{fig:exp_comparison}
\vspace*{-6pt}
\end{figure*}

\subsection{GAN inversion for textures}

Once a reliable generator is obtained, texture analysis can be performed via GAN inversion. Inverting a latent space trained on textures poses distinct challenges. Most generative models produce images with well-defined objects of interest, \eg, a face or bed. However, for textures, everything is of interest and the stochastic placement of textons renders commonly used location-wise losses, such as L2 loss and content loss \cite{Johnson:Perceptual:ECCV2016}, inapplicable. Instead, we use the style loss by Gatys \etal \cite{ Gatys:TextureSynthCNN:NIPS2015} who extract the Gram matrices of feature maps from a pretrained VGG-16 network $\Phi$\cite{Simonyan:VGG:ICLR2015}. The process of searching for optimal latent code $\mathbf{z}^*$ for any texture $\Omega_i$ is given via:
\begin{align}
&\mathcal{L}(\mathbf{I}_1,\mathbf{I}_2) = \sum_{\mathnormal{l}}[\frac{\Phi_{\mathnormal{l}}(\mathbf{I}_1)\Phi_{\mathnormal{l}}(\mathbf{I}_1)^T-\Phi_{\mathnormal{l}}(\mathbf{I}_2)\Phi_{\mathnormal{l}}(\mathbf{I}_2)^T}{C_{\mathnormal{l}}\times N_{\mathnormal{l}}^2}]^2\\
&\mathbf{\mathbf z}^* = \arg\min_{\mathbf{z}}\mathbb{E}_{(\mathbf{n}, \Delta_G,\mathbf{I}\sim\Omega_i)}[ \mathcal{L}(G_\mathbf{\mathbf\theta}(\mathbf{z}, \mathbf{n}, \Delta_G),\;\mathbf{I})] 
\label{eq:gan_invert}
\end{align}
where $\mathnormal{l}$ is layer index, and $\Phi_{\mathnormal{l}}$, $N_{\mathnormal{l}}$, $C_{\mathnormal{l}}$ represent feature maps, spatial size and channel at $\mathnormal{l}$-th layer of $\Phi$, respectively. Note that a common practice in the literature of GAN inversion is to use $\mathbf{w}$ or $\mathbf{w}^+$ as the optimization variable instead of the raw latent code $\mathbf{z}$, where $\mathbf{w}$ comes from the style mapping network with $\mathbf{z}$ as input, and $\mathbf{w}^+$ is the aggregation of all $\mathbf{w}$ across different layers.
\begin{table}[t]
\centering
\begin{tabular}{||l | c||} 
 \hline
  & FID $\downarrow$ \\ [0.5ex] 
 \hline\hline
 PSGAN\cite{Bergmann:PSGan:ICML2017} & 133.72$\pm$2.46   \\ 
 \hline
 StyleGAN-2\cite{Karras:StyleGAN2:CVPR2020} & 72.48$\pm$1.86  \\
 \hline
 Proposed & $\mathbf{70.05\pm1.41}$ \\
 \hline
 \hline
 Training set & 1.58$\pm1.10$ \\
 \hline
\end{tabular}
\vspace*{-5pt}
\caption{Quantitative evaluation of different methods}
\label{fig:fid}
\vspace*{-20pt}
\end{table}

\section{Experimental results}
\begin{figure*}[t]
\begin{subfigure}[b]{1\textwidth}
\centering
\includegraphics[width=0.12\textwidth]{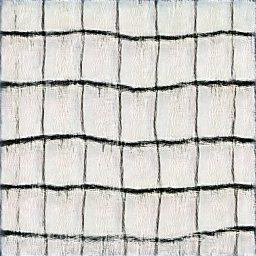} 
\includegraphics[width=0.12\textwidth]{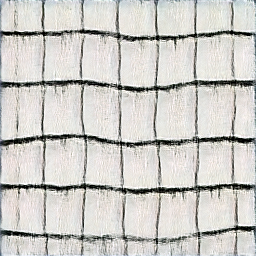} 
\includegraphics[width=0.12\textwidth]{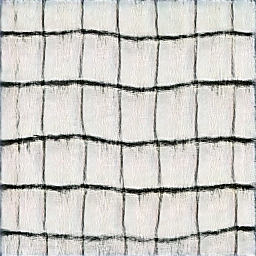}
\includegraphics[width=0.12\textwidth]{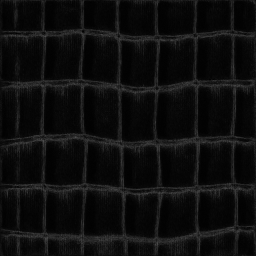}
\includegraphics[width=0.12\textwidth]{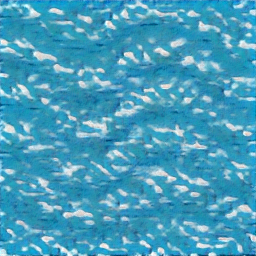}
\includegraphics[width=0.12\textwidth]{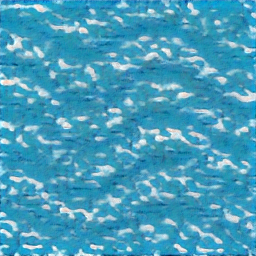}
\includegraphics[width=0.12\textwidth]{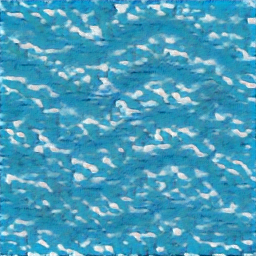}
\includegraphics[width=0.12\textwidth]{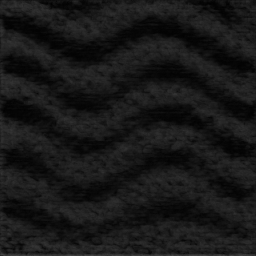}
\caption{StyleGAN-2: Exhibits intra-texture mode collapse}
\label{fig:exp_intra_mc_stylegan2}
\end{subfigure}

\begin{subfigure}[b]{1\textwidth}
\centering
\includegraphics[width=0.12\textwidth]{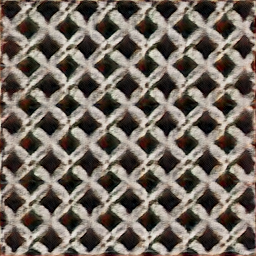}
\includegraphics[width=0.12\textwidth]{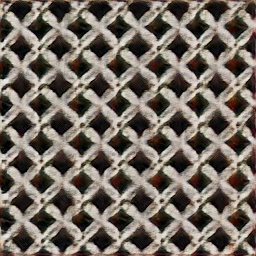}
\includegraphics[width=0.12\textwidth]{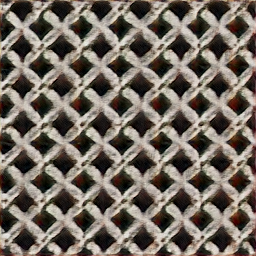}
\includegraphics[width=0.12\textwidth]{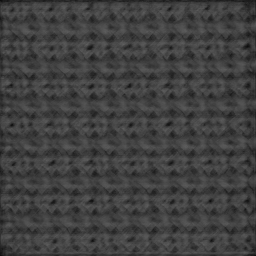}
\includegraphics[width=0.12\textwidth]{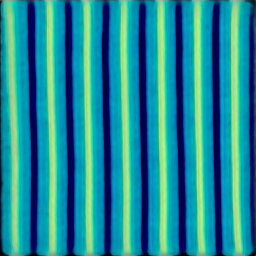}
\includegraphics[width=0.12\textwidth]{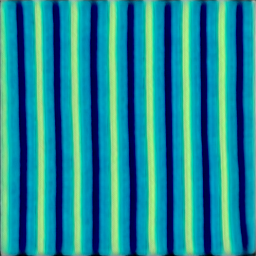}
\includegraphics[width=0.12\textwidth]{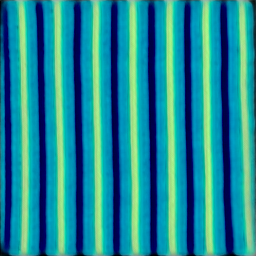}
\includegraphics[width=0.12\textwidth]{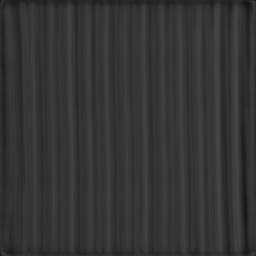}
\caption{Proposed approach: No intra-texture mode collapse}
\label{fig:exp_intra_mc_ours}
\end{subfigure}
\vspace*{-14pt}
\caption{Different crops sampled from the same latent vector, and their associated standard deviation map $\sigma_{\mathbf{z}}$ on the right. 
Higher intensity in the $\sigma_{\mathbf{z}}$ map indicates lower likelihood of anchoring artifacts, while darker regions indicate severe intra-texture mode collapse.}
\label{fig:exp_intra_mode_collapse}
\vspace*{-6pt}
\end{figure*}
\label{sec:experiment}
\subsection{Dataset construction}
There exist multiple texture datasets (\eg\cite{Cimpoi:DTD:CVPR2014}) in the literature. Unfortunately, most of them either lack inter-texture diversity or sufficient spatial resolution for intra-texture distribution. To address this limitation, we collect a more comprehensive dataset of textures. We include a texture image into the dataset if it fits the following criteria:
(1) It is perceptually uniform;
(2) it contains sufficient independent $256\times256$ crops to learn the intra-texture distribution; 
(3) each $256\times256$ crop contains enough texton repetitions (at least 5 in each dimension) to form a texture; and
(4) it is under either Creative Commons Public Domain license (CC0) or custom website license for free academic use. 
After extensive search on multiple stock image websites, we obtain 500 quality texture images, ranging from natural to artificial, periodic to stochastic, and fine-grained to coarse, and each image contains roughly 20 to 50 independent crops.

\subsection{Training settings}
In our experiment, we uniformly sampled 2 crops of $256\times256$ from each texture in a mini-batch of 8 distinct textures, to explicitly enforce the learning of intra-texture distribution on the discriminator network $D_{\boldsymbol\phi}$. The same trick can be applied to the generator network $G_{\boldsymbol\theta}$ by feeding multiple noise samples $\mathbf{n}$ conditioned on the same latent variable $\mathbf{z}$, but this resulted in a slower convergence of the generator and no significant performance gain. We adopted the Wasserstein distance in (\ref{eq:loss_D}) and (\ref{eq:loss_G}) as the losses with gradient penalty $=0.01$ to impose Lipschitz continuity, and the discriminator parameters $\boldsymbol\phi$ were updated twice, followed by one generator update ($3\times10^5$ generator iterations in total). We disabled the mixing regularization and the path-length regularization as they are time-consuming. For other hyperparameters, we followed the default protocols of StyleGAN-2, including latent space dimensionality $D=512$, learning rate $=0.002$, Adam optimizer, and exponential moving average of $G_{\boldsymbol\theta}$.


Regarding the settings of TB, each module has $P=16$ learnable texton vectors. We applied the TB module to all layers of spatial resolution up to $64\times64$ to prevent high-frequency artifacts. The spatial size of the bottom TB module was set to $4\times4$ during training, and the number of channels of each texton vector was fixed at C$=512$. The spatial size as well as channel size of all remaining TB modules were designed to match the feature maps of their preceding blocks of Styled Conv as shown in Fig.~\ref{fig:msTextureGen}. At test time, the final output image resolution can be varied by simply modifying the spatial size of the bottom module.

\subsection{Comparisons}
\begin{figure*}[t]
\centering
\begin{subfigure}[b]{1\textwidth}
\includegraphics[width=0.12\textwidth]{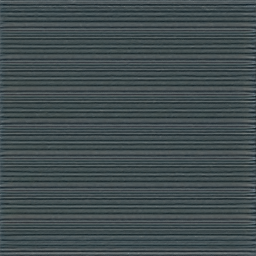} 
\includegraphics[width=0.12\textwidth]{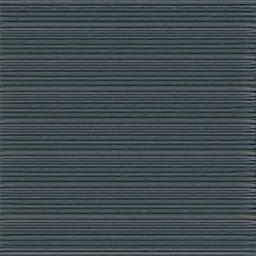}
\includegraphics[width=0.12\textwidth]{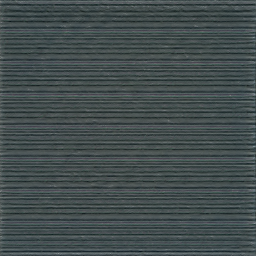}
\includegraphics[width=0.12\textwidth]{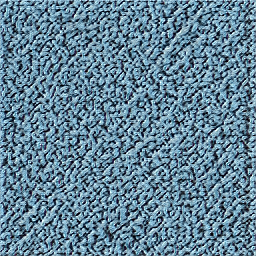} 
\includegraphics[width=0.12\textwidth]{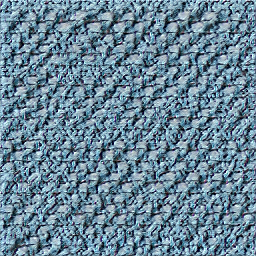}
\includegraphics[width=0.12\textwidth]{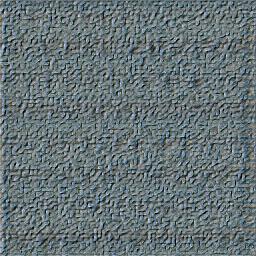} 
\includegraphics[width=0.12\textwidth]{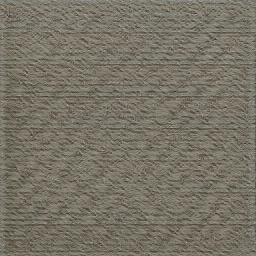}
\includegraphics[width=0.12\textwidth]{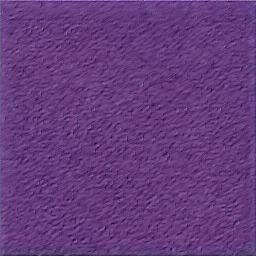}
\caption{Original training strategy of StyleGAN-2, \ie non-saturating loss, shows inter-texture mode collapse}
\label{fig:exp_inter_mc_original}
\end{subfigure}


\begin{subfigure}[b]{1\textwidth}
\includegraphics[width=0.12\textwidth]{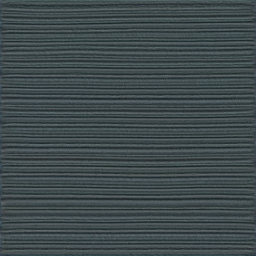}
\includegraphics[width=0.12\textwidth]{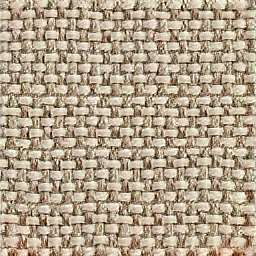}
\includegraphics[width=0.12\textwidth]{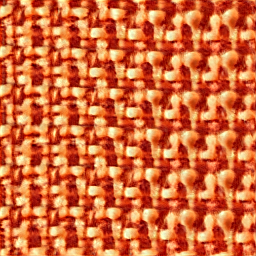}
\includegraphics[width=0.12\textwidth]{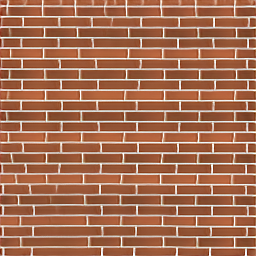}
\includegraphics[width=0.12\textwidth]{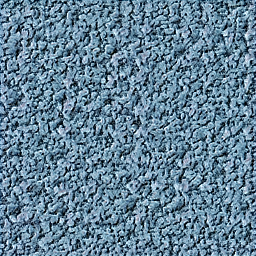} 
\includegraphics[width=0.12\textwidth]{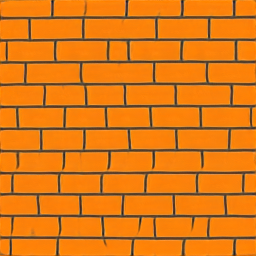}
\includegraphics[width=0.12\textwidth]{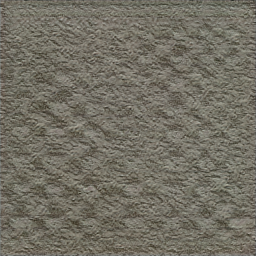} 
\includegraphics[width=0.12\textwidth]{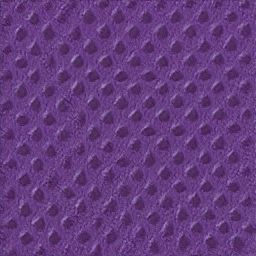}
 \caption{StyleGAN-2 with Wasserstein distance + discriminator noise}
\end{subfigure}
\vspace*{-14pt}
\caption{Inter-texture mode collapse experiments}
\label{fig:exp_inter_mode_collapse}
\vspace*{-10pt}
\end{figure*}
We compare our approach with PSGAN, which also aims to model periodic textures, and the baseline StyleGAN-2. All methods were trained on our dataset, and we provide a list of textures sampled from each method. As shown in Fig.~\ref{fig:exp_comparison}, the proposed module outperforms PSGAN and StyleGAN-2 in terms of diversity and image quality. We will provide more samples in the supplementary material. We also evaluate the FID \cite{Heusel:FID:NIPS2017} for each model in Table~\ref{fig:fid}.
The distribution of training set was computed by sampling 20 crops from each texture. For each method, 1000 latent codes were sampled, and 20 different texture crops were obtained for each code. Each method was repeated 5 times, and the 95\% confidence intervals are also provided.

\vspace*{-10pt}
\paragraph{Intra-texture mode collapse.} Figure~\ref{fig:exp_intra_mode_collapse} shows that the proposed approach yields a significant improvement in terms of intra-texture diversity, that is, the reconstructed textures do not all correspond to the same crop.
The figure also shows the map of pixel-wise standard deviations  $\sigma_{\mathbf{z}}\in\mathbb{R}^{H\times W}$ conditioned on $\mathbf{z}$ defined by:
\begin{equation}
\sigma_{\mathbf{z}}=\sqrt{\mathbb{E}_{\mathbf{n} | \mathbf{z}}[(G_\mathbf{\mathbf\theta}(\mathbf{z}, \mathbf{n})-\mathbb{E}_{\mathbf{n} | \mathbf{z}}[G_\mathbf{\mathbf\theta}(\mathbf{z}, \mathbf{n})])^2]}\label{eq:std_map}
\end{equation}  
Pixels with higher intensities in the $\sigma_{\mathbf{z}}$ map are less likely to suffer from anchoring artifacts, while darker regions in the map indicate severe intra-texture mode collapse.

To quantify the intra-texture mode collapse of a synthesized texture, we now introduce a novel and intuitive measure we call {\it thresholded invariant pixel percentage (TIPP)}, which is calculated as follows: 
\begin{equation}
\text{TIPP}_t(\mathbf{z})=\frac{1}{H\times W}\sum_h^H\sum_w^W\1(\sigma_{\mathbf{z}}[h, w]\leq t)\label{eq:tipp}
\end{equation} 
where $\sigma_{\mathbf{z}}\in\mathbb{R}^{H\times W}$ is the pixel-wise standard deviation map defined in (\ref{eq:std_map}), $t$ is the threshold, and $\1(\cdot)$ is an indicator function that returns 1 if the condition in parenthesis is met and 0 otherwise. 
Intuitively, $\text{TIPP}_t$ calculates the percentage of pixels with $\sigma_{\mathbf{z}}<t$. Pixels with low standard deviation have a strong invariance, and thus the higher the $\text{TIPP}$ value the worse the intra-texture mode collapse. 

\begin{figure}[t]
\begin{center}
\includegraphics[width=0.35\textwidth]{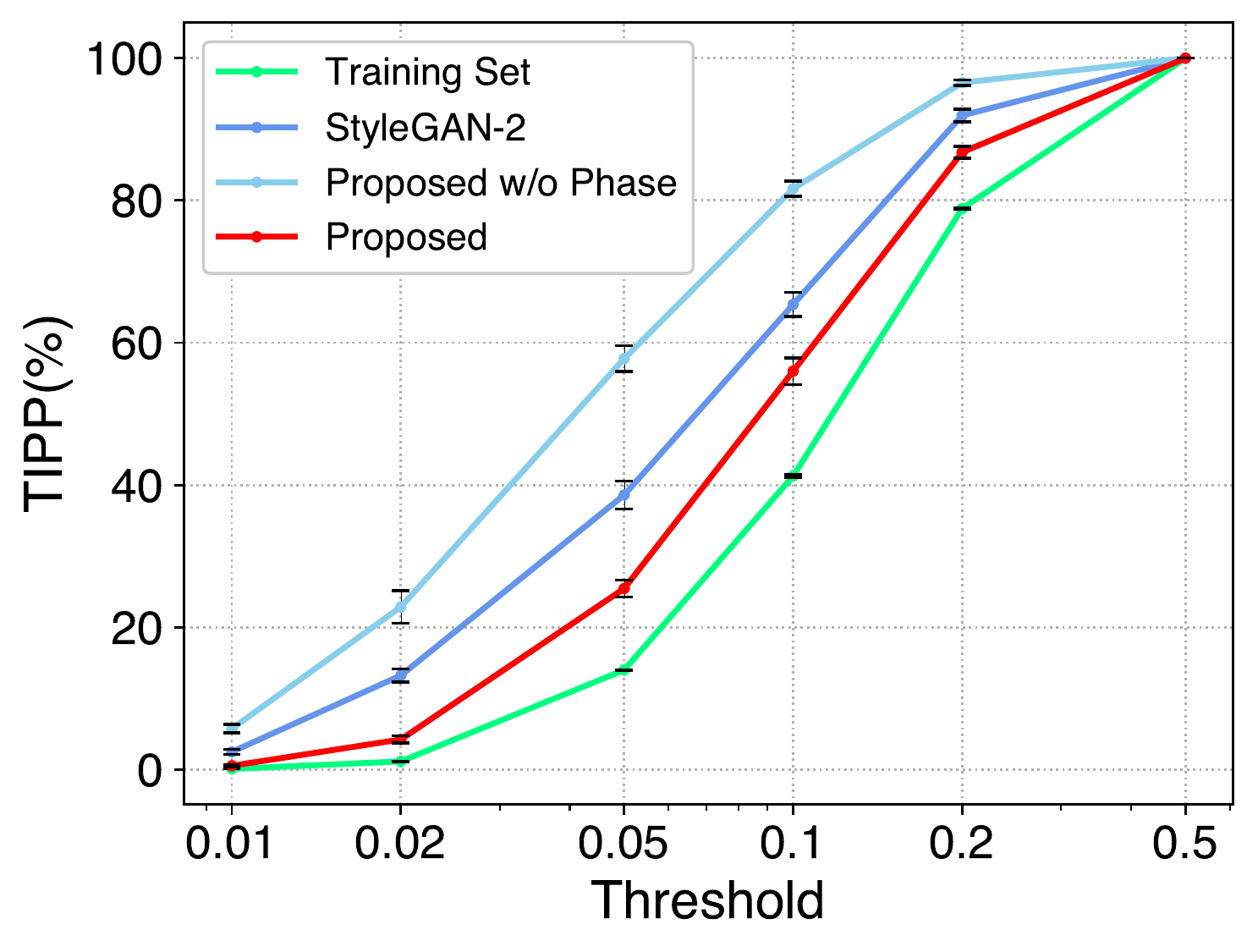}
\end{center}
\vspace*{-12pt}
\caption{TIPP measure of intra-texture mode collapse shows that proposed method outperforms StyleGAN-2, and disabling random phase $\Delta$ causes significant drop in TIPP.}
\label{fig:no_phase}
\vspace*{-8pt}
\end{figure}

To evaluate each model, we sampled 1000 latent codes $\mathbf{z}$, synthesized 20 crops per $\mathbf{z}$, and calculated TIPP averaged over the codes $\mathbf{z}$ for different thresholds.
TIPP(\%) can also be calculated for the training set, where the averaging is over crops rather than latent codes.  For that, we sampled 20 crops from each texture in the training set. 
Figure~\ref{fig:no_phase} shows that the proposed method consistently outperforms StyleGAN-2 in terms of intra-texture mode collapse.

\begin{figure}[t]   
\centering
\begin{subfigure}[t]{0.115\textwidth}
\centering
\includegraphics[width=1\textwidth]{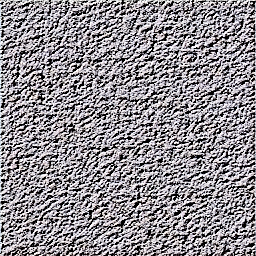} 
\caption{Ground truth}
\end{subfigure}
\begin{subfigure}[t]{0.115\textwidth}
\centering
\includegraphics[width=1\textwidth]{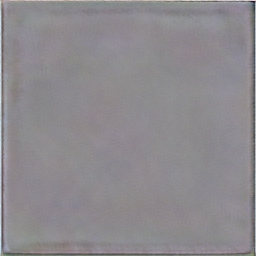} 
\caption{L2 loss}
\end{subfigure}
\begin{subfigure}[t]{0.115\textwidth}
\centering
\includegraphics[width=1\textwidth]{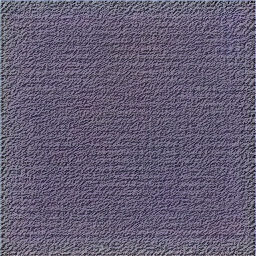} 
\caption{Content loss}
\end{subfigure}
\begin{subfigure}[t]{0.115\textwidth}
\centering
\includegraphics[width=1\textwidth]{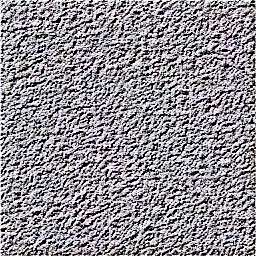} 
\caption{Gram matrices}
\end{subfigure}
\caption{Loss functions for texture GAN inversion}
\label{fig:exp_gan_invert}
\vspace*{-10pt}
\end{figure}
\vspace*{-10pt}

\paragraph{Inter-texture mode collapse.}
We designed an experiment to investigate inter-texture mode collapse. We selected 8 textures of distinguishable properties, and independently sampled 8 fixed latent codes, from which the generator samples during training. Our expectation was that StyleGAN-2 would overfit the data, allocating each latent code to a different texture, which would indicate strong disentanglement between latent code and noise injection. To our surprise, as shown in Fig.~\ref{fig:exp_inter_mc_original}, the original StyleGAN-2 strategy with non-saturating loss performs poorly on this small set. We then applied the Wasserstein distance combined with adding Gaussian noise ($\sigma=0.01$) to the discriminator input and found that it improves diversity as shown in Fig.~\ref{fig:exp_inter_mode_collapse}. Therefore, we adopted the Wasserstein distance and discriminator noise as the default training configuration for both the proposed method and StyleGAN-2 (included in the results shown in Fig.~\ref{fig:exp_comparison}).

\begin{figure*}[t]
\centering
\begin{subfigure}[b]{1\textwidth}
\includegraphics[width=0.12\textwidth]{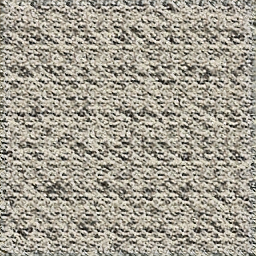}
\includegraphics[width=0.12\textwidth]{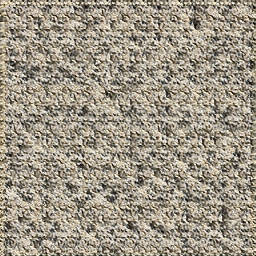}
\includegraphics[width=0.12\textwidth]{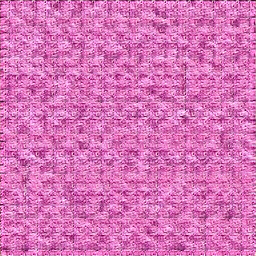}
\includegraphics[width=0.12\textwidth]{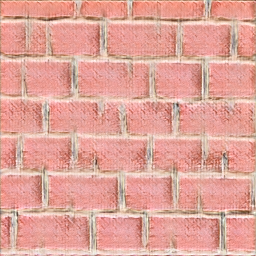}
\includegraphics[width=0.12\textwidth]{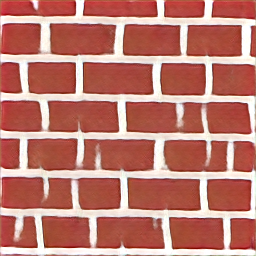} 
\includegraphics[width=0.12\textwidth]{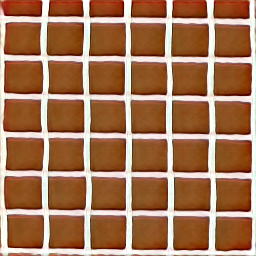}
\includegraphics[width=0.12\textwidth]{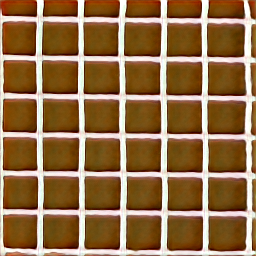} 
\includegraphics[width=0.12\textwidth]{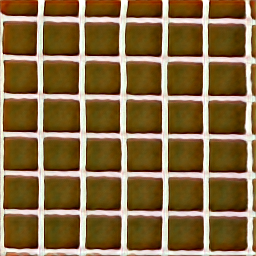}
 \caption{Traversing in proposed $\mathbf{Z}$ latent space}
\end{subfigure}
\begin{subfigure}[b]{1\textwidth}
\includegraphics[width=0.12\textwidth]{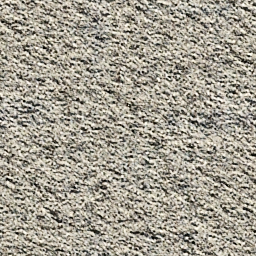} 
\includegraphics[width=0.12\textwidth]{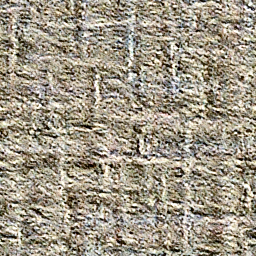} 
\includegraphics[width=0.12\textwidth]{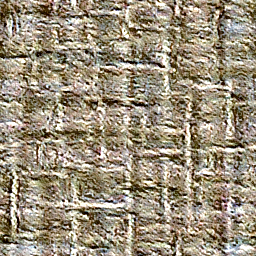} 
\includegraphics[width=0.12\textwidth]{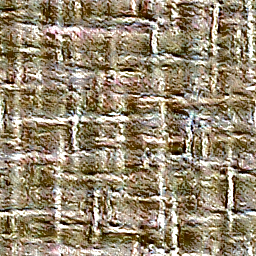} 
\includegraphics[width=0.12\textwidth]{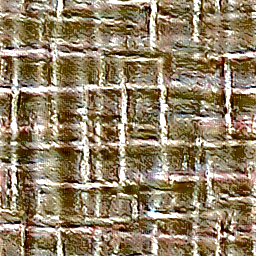} 
\includegraphics[width=0.12\textwidth]{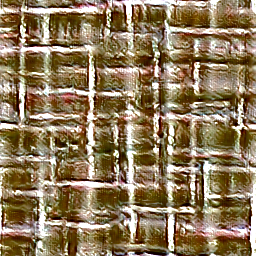}
\includegraphics[width=0.12\textwidth]{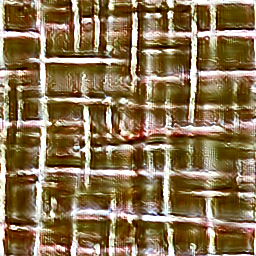} 
\includegraphics[width=0.12\textwidth]{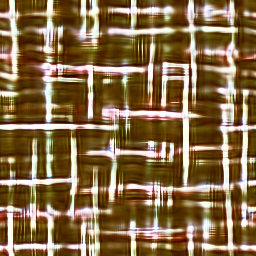} 
\caption{Based on Portilla and Simoncelli parametrization\cite{portilla_simoncelli}}
\label{fig:interpolate_by_PS}
\end{subfigure}
\vspace*{-14pt}
\caption{Comparison between texture interpolation based on proposed latent space and Portilla and Simoncelli parametrization.}
\label{fig:exp_latent_interpolate}
\vspace*{-6pt}
\end{figure*}

\subsection{GAN inversion}
In this experiment, we used $relu1\_2$, $relu2\_2$, $relu3\_3$ and $relu4\_3$ layers to extract feature maps and compute Gram matrices via (\ref{eq:gan_invert}). For completeness, we also investigated the efficacy of L2 loss and content loss \cite{Johnson:Perceptual:ECCV2016}. For all losses, the inversion was performed on the same optimization variable $\mathbf{w}$. We avoided the use of $\mathbf{w}^+$ as we prefer a unified global representation for textures. Other shared hyperparameters include learning rate of 0.001, a total of 5000 iterations, and Adam optimizer with default settings. Results are shown in Fig.~\ref{fig:exp_gan_invert} and are consistent with our expectations, as the characterization of a texture should minimize location-wise correspondence due to its stochastic nature.

\subsection{Latent space interpolation}

Interpolation in the latent space can be performed by traversing in the $\mathbf{Z}$ or $\mathbf{W}$ space, which is mapped from input $\mathbf{z}$ via a MLP. In both learned latent spaces, there is a gradual transition from one texture to the other via intermediate uniform textures, as shown in Fig.~\ref{fig:exp_latent_interpolate}. In contrast, the interpolations produced by the classic Portilla and Simoncelli \cite{portilla00} method consist of a mixture of the endpoint images rather than homogeneous intermediate textures.

\subsection{Ablation studies}
\paragraph{Is random phase noise $\Delta$ needed?}
To demonstrate the necessity of $\Delta$, we re-trained the model with fixed phase. As expected, the module degenerates to another form of spatial anchoring, severely damaging the intra-texture diversity. As shown in Fig.~\ref{fig:no_phase}, such setting degrades TIPP by a considerable margin.
\vspace*{-8pt}

\begin{figure}[t]
\begin{subfigure}[b]{0.2366\textwidth}
\includegraphics[width=0.325\textwidth]{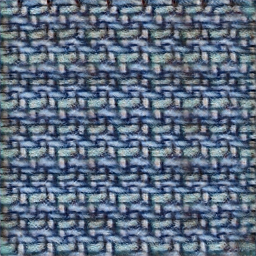} 
\includegraphics[width=0.65\textwidth]{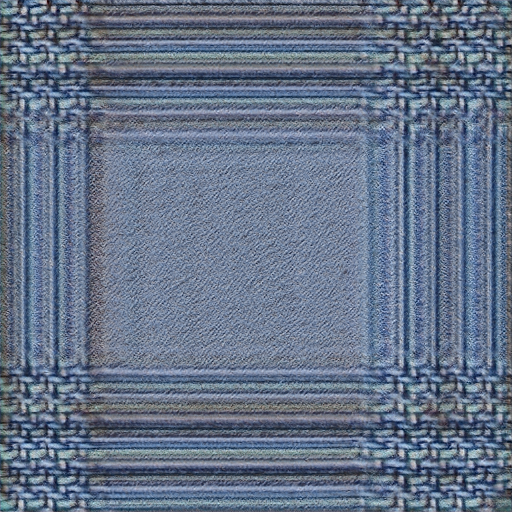} 
\caption{TB Module Only at the Bottom}
\label{fig:exp_no_ms}
\end{subfigure}
\begin{subfigure}[b]{0.2366\textwidth}
\includegraphics[width=0.325\textwidth]{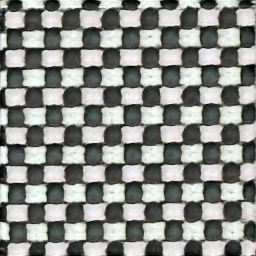}
\includegraphics[width=0.65\textwidth]{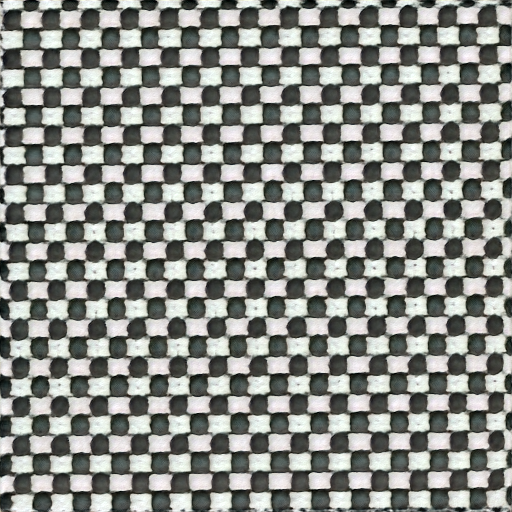}
\caption{Multi-scale TB Modules}
\label{fig:exp_ms_tb}
\end{subfigure}
\caption{Multi-scale texton broadcasting experiments}
\label{fig:exp_ms_tb_or_not}
\vspace*{-8pt}
\end{figure}

\paragraph{Multi-Scale texton broadcasting.}
We demonstrated the importance of multi-scale texton broadcasting by removing all but the bottom TB modules. Such an ablated model is capable of generating quality textures with the same size as the training images. However, when generalized to arbitrary sizes, only the 4 corners, shown in Fig.~\ref{fig:exp_no_ms}, resemble their low-resolution counterpart and the model fails to render the central area. We attribute this to the zero-padding at bottom layers, consistent with the study in \cite{Xu:PositionEncode:CVPR2021} where zero-padding is shown to have an implicit encoding of location. By introducing our module in a multi-scale fashion, the influence of zero-padding can be substantially reduced. 
\vspace*{-10pt}

\paragraph{Mapping latent code to TB via a MLP?}
We also conducted experiments with the trainable parameters in the TB module linked to the latent code as in PSGAN via a MLP\cite{Bergmann:PSGan:ICML2017}, \eg, $f_i=MLP_{f_i}(\textbf{z})$. We empirically found that such a model is unstable to train and struggles to converge. We hypothesize that if the module parameters are conditioned on the latent code, then the entanglement between location and latent space is aggravated because the resulting broadcast maps are directly affected by the latent code, and the latent space no longer works in a channel-wise manner, contrary to the design philosophy of StyleGAN. 

\section{Limitations}
While the proposed approach yields promising results,  understanding textures remains a difficult task as such a powerful model still fails to encompass a training set of 500 textures. The immense diversity across textures as well as within textures poses unique challenges, which result in a relatively high FID compared with images of other domain. For texture analysis, optimization-based iterative inversion is less time-efficient compared with direct statistics computation, and systematic procedures of exploring the learned latent space is still pending. In future work, we aim to expand our dataset as well as address fundamental limitations of our model for a more thorough study.
\vspace*{-8pt}
\section{Conclusions}
We performed an in-depth analysis of StyleGAN-2, utilized the noise injection as a means of modeling intra-texture distribution, recognized its shortcoming of spatial anchoring artifacts, and designed a model that has an architectural inductive bias more aligned with textures. To demonstrate the effectiveness of our module, we created a high-resolution dataset, proposed a novel measure (TIFF) that quantifies the anchoring artifacts, and conducted extensive experiments to both quantitatively and qualitatively evaluate the proposed modules. The proposed work facilitates universal textures synthesis and enables a potentially unified formulation of texture analysis and synthesis with help of deep neural networks.

{\small
\bibliographystyle{ieee_fullname}
\bibliography{IEEEabrv, egbib, cvision, retrieval, texture, coding, JUELIN}
}

\end{document}